%% file: main.tex
\documentclass[onecolumn]{IEEEtran}

\usepackage[utf8]{inputenc} 
\usepackage[T1]{fontenc}
\usepackage{url}              

\usepackage[cmex10]{amsmath}  
\usepackage{mleftright}       
\mleftright                   

\usepackage{graphicx}         
\usepackage{booktabs}         

\usepackage[utf8]{inputenc}
\usepackage[T1]{fontenc}    
\usepackage{url}            
\usepackage{booktabs}       
\usepackage{nicefrac}       
\usepackage{microtype}      
\usepackage{xcolor}         
\usepackage{xspace}         
\usepackage{amsmath, amssymb}
\usepackage{amsmath,amssymb,amsfonts}

\usepackage{algorithm}

\newtheorem{theorem}{Theorem}

\newtheorem{corollary}{Corollary}

\usepackage{algorithm}
\usepackage{algpseudocode}

\usepackage{cite}

\begin{document}

\title{Learning for Bandits under Action Erasures} 
\date{}
\author{\IEEEauthorblockN{Osama A. Hanna$^\dagger$$^*$, Merve Karakas$^\dagger$$^*$, Lin F. Yang$^\dagger$ and Christina Fragouli$^\dagger$\\ 
$^\dagger$University of California, Los Angeles\\
Email:\{ohanna, mervekarakas, linyang, christina.fragouli\}@ucla.edu}
}
\maketitle

\allowdisplaybreaks
\begin{abstract}
    \input{abstract}
\end{abstract}

\input{content}

\bibliographystyle{IEEEtran}
\bibliography{references}

\end{document}

%% file: abstract.tex
We consider a novel multi-arm bandit (MAB) setup,  where  a learner needs to communicate the actions to  distributed agents over erasure channels, while the rewards for the actions  are directly available to the learner through external sensors. 
In our model, while the distributed  agents know if an action is erased,  the central learner does not (there is no feedback), and thus does not know whether the observed reward resulted from the desired action or not. 
We propose a scheme that can work on top of any (existing or future) MAB algorithm and make it robust to action erasures. Our scheme results in a worst-case regret over  action-erasure channels that is at most a factor of $O(1/\sqrt{1-\epsilon})$ away from the no-erasure worst-case regret of the underlying MAB algorithm, where $\epsilon$ is the erasure probability.
We  also propose a modification of the successive arm elimination algorithm and prove that its worst-case regret is $\Tilde{O}(\sqrt{KT}+K/(1-\epsilon))$, which we prove is  optimal by providing a matching  lower bound.

%% file: content.tex
\section{Introduction}

Multi-armed bandit (MAB) problems have gained popularity in a wide range of applications that include recommendation systems, clinical trials, advertising, and distributed robotics \cite{HebertRobotic, BouneffoufMABsurvey}. MABs are sequential decision problems where a learner, at each round, selects an action from a set of $K$ available actions (arms) and receives an associated reward; the  aim is to maximize the total reward over $T$ rounds. 
In this paper, we develop a theoretical framework that explores a new setup, MAB systems with action erasures. 

In particular, we consider a MAB system where the learner needs to communicate the actions to be taken to  distributed agents over wireless channels subject to erasures  with some probability $\epsilon$, while the rewards for the actions taken are directly available to the learner through external sensors. 
For example, the learner may be regulating drone traffic in a slowly varying environment (weather conditions, obstacles) where passing by drones receive which path to take or which maneuvers to perform, and the learner monitors the outcome
through video cameras, magnetic and other sensors. Similarly, since online learning has been used in medical micro-robots \cite{Yang22microrobots, Zou22microswimmers},  our setup can help with problems such as directing inside veins micro-robots on how to move or how much of a medical substance to release, and observing the results through patient medical imaging and vital sign measurements.  
In our model, while the agent knows if an action is erased,  the learner does not (there is no feedback); therefore, the learner at each round receives a reward that may be a result of playing the  requested action or another action (if an erasure occurs).

Note that at any given round, although the requested action may be erased and not received by the agent, the agent still needs to play an action -  the micro-robot needs to release some amount (perhaps zero) of substance and the drone needs to continue moving in some way. 
We make the assumption that if at a round the agent does not receive an action, the agent continues to play the  most recently received action, until a new action is successfully received to replace it. This is we believe  a reasonable assumption. Indeed, during exploitation, we would like the agents to persistently play the identified optimal action, which this strategy achieves. In contrast, as we discuss in Section~\ref{sec:prelim}
playing a random  or a fixed action results in suboptimal performance, while using more sophisticated strategies (e.g., keeping track of all previous actions played and accordingly optimizing) may not be possible in distributed systems, where multiple agents may be playing different actions and where agents may not even be able to observe the rewards.

To the best of our knowledge, this setup has not been systematically studied, although there exists extensive MAB literature.

Indeed, existing work has  examined 
MAB systems under a multitude of setups and constraints such as stochastic bandits \cite{Auer2002, Thompson1933ONTL, Lai1987AdaptiveTA}, adversarial bandits \cite{Auer95Adv, pmlr-v23-bubeck12b, Auer2002b}, or contextual bandits \cite{Auer2002b, Li2010Contextual}; and over a number of distributed setups as well \cite{ShahMABMultiAgent, KalathilDecentralMultiPMA, LANDGREN21DistMABMultiAgent, Landgren2019DistributedMM}. In particular, a number of works examine the case where rewards are corrupted adversarially or stochastically \cite{Lykouris18,pmlr-v99-gupta19a, Kapoor19ProbCorReward, amir2020prediction}. However,  all the works we know of assume that the requested actions can be sent through perfect communication channels. This  assumption makes sense for delay-tolerant applications, where we can leverage feedback and error-correcting codes to ensure perfect communication of the desired action; yet as the applicability of the MAB framework expands to agents, such as micro-robots, that are communication limited, we believe that our proposed model can be of interest, both in terms of theory and practice.

The main questions we ask (and answer) in this paper are: for our action erasure model (i) what is an optimal strategy? and (ii) if a MAB algorithm is already deployed, can we couple it, at low regret cost, with a generic layer, akin to erasure coding, that makes the MAB algorithm operation robust to erasures?
 
Our main contributions include:
\begin{itemize}
\item We propose a scheme that can work on top of any MAB algorithm and make it robust to action erasures. Our scheme results in a worst-case regret over the erasure channel that is at most a factor of $O(1/\sqrt{1-\epsilon})$ away from the no-erasure worst-case regret of the underlying MAB algorithm, where $\epsilon\in[0,1)$ is the erasure probability.
\item We propose a modification of the successive arm elimination algorithm and prove that its worst-case regret is $\Tilde{O}(\sqrt{KT}+K/(1-\epsilon))$. 
\item  We prove a matching $\Omega(\frac{K}{1-\epsilon})$ regret lower bound.
\end{itemize}
\subsection{Related Work}

{\bf MAB Algorithms.}
{Over the years, several stochastic MAB algorithms that achieve optimal or near-optimal regret bounds have been proposed under different assumptions for the environment or model parameters. 

For instance, over a horizon of length $T$, Thompson sampling \cite{Thompson1933ONTL} and UCB \cite{Auer2002, Lai1987AdaptiveTA}, achieve a gap dependent regret bound of $\tilde{O}(\sum_{i:\Delta_i>0}\frac{1}{\Delta_i})$ and a worst-case regret bound of $O(\sqrt{KT\log{T}})$, where $\tilde{O}$ hides $\log$ factors. These algorithms achieve nearly optimal regret; a lower bound of $\Omega(\sum_{i:\Delta_i>0}\frac{1}{\Delta_i})$ on the gap-dependent regret was proved in \cite{lai1985asymptotically} while a lower bound on the worst-case regret of $\Omega(\sqrt{KT})$ was proved in \cite{auer1995gambling}. However, such algorithms are not robust to action erasures.}\\
{\bf MAB with Adversarial Corruption.} 
Recently, \cite{Lykouris18} introduced stochastic multi-armed bandits with adversarial corruption, where the rewards are corrupted by an adversary with a corruption budget $C$, an upper bound on the accumulated corruption magnitudes over $T$ rounds. \cite{Lykouris18} proposes a simple elimination algorithm that achieves $\Tilde{O}(C\sqrt{KT})$ regret bound in the worst-case. Later, \cite{pmlr-v99-gupta19a} improved the dependency on $C$ to be additive by achieving a worst-case regret bound of $\Tilde{O}(\sqrt{KT}+KC)$ while also providing a regret lower bound of $\Omega(C)$. The work in \cite{amir2020prediction} further improves the worst-case regret bound to $\Tilde{O}(\sqrt{KT}+C)$ when the optimal arm is unique. The work in \cite{Kapoor19ProbCorReward} studies a similar problem where rewards are corrupted in each round independently with a fixed probability. \cite{Kapoor19ProbCorReward} achieves a regret of $\Tilde{O}(\sqrt{KT} + CT)$, where $C$ is an upper bound on the expected corruption, and proves a $\Omega(CT)$ lower bound for their model.

While our model can be reduced to MABs with reward corruption, the amount of corruption $C$ amounts to $\Omega(\epsilon T)$. This causes the best-known algorithms for MABs with adversarial corruption to suffer a linear regret, given the increased exploration required for the large amount of corruption.  {In contrast, by exploiting the fact that corruptions are in actions, we achieve a gap-dependent regret bound of $\tilde{O}(\sum_{i:\Delta_i>0}\frac{1}{\Delta_i})$ and a worst-case regret bound of $\tilde{O}(\sqrt{KT}+\frac{K}{1-\epsilon})$. 

\subsection{Paper Organization} Section~\ref{sec:model} introduces our system model and notation; Section~\ref{sec:prelim} discusses
some straightforward approaches and why they would fail; Section~\ref{sec:alg1}, and Section~\ref{sec:alg2} describe our proposed algorithms and upper bounds; and Section~\ref{sec:lower-bnd} provides our lower bound.

\section{Problem Formulation}
\label{sec:model}
\noindent{\bf Standard MAB Setup.} We consider a MAB problem over a horizon of length $T$, where a learner interacts with an environment by pulling arms from a set of $K$ arms (or actions). That is, we assume operation in $T$ discrete times, where at each time $t\in [T]$, the learner sends the index of an arm $a_t$ to an agent; the agent pulls the arm $a_t$ resulting in a reward $r_t$ that can be observed by the learner. The learner selects the arm $a_t$ based on the history of pulled arms and previously seen rewards $a_1,r_1,...,a_{t-1},r_{t-1}$. The reward $r_t$ is sampled from an unknown distribution with an unknown mean $\mu_{a_t}$. The set of $K$ distributions for all arm rewards is referred to as a bandit instance.
{We use $\Delta_i$ to denote the gap between the mean value of arm $i$ and the best (highest mean) arm.} {For simplicity, we follow the standard assumption that rewards are supported on $[0,1]$. The analysis directly extends to subGaussian reward distributions.}

\noindent{\bf MAB with Action Erasures.} 
We assume that the learner is connected to the agent over an erasure channel, where the action index sent by the learner to the agent at each time $t$ can be erased with probability $\epsilon \in [0, 1)$. We follow the standard erasure channel model where erasures at different times are independent. We assume no feedback in the channel; in particular, while the agent knows when the action is erased (does not receive anything at time $t$), the learner does not. 
Moreover, we assume that the learner can directly observe the reward of the action played $r_t$.

\noindent{\bf Agent Operation.} Recall that  at each time $t$, the learner transmits an action index $a_t$. The agent plays the most recent action she received: in particular, at time $t$, the agent plays the action $\tilde{a}_t=a_t$ if no erasure occurs, while if
an erasure occurs, $\Tilde{a}_t=\Tilde{a}_{t-1}$. We assume that $\tilde{a}_0$ is initialized uniformly at random, i.e., if the first action is erased, the agent chooses $\tilde{a}_1$ uniformly at random. In the example described in Table~\ref{table:erasure_ex}, $\Tilde{a}_4=\Tilde{a}_3=\Tilde{a}_2=a_2$. {Section~\ref{sec:prelim} motivates why we select this particular agent operation, by arguing that alternative simple strategies can significantly deteriorate the performance.}

\begin{table}[t!]
\centering
\caption{Example of actions played under action erasures.}
\begin{tabular}{|c| c c c c c c |}
\hline
Round ($t$) & \textbf{1} & \textbf{2} & \textbf{3} & \textbf{4} & \textbf{5} & \textbf{...} \\
\hline
Learner ($a_t$) & 1 & 3 & 2 & 4 & 2 & ... \\
Agent ($\hat{a}_t$) & 1 & 3 & 3 & 3 & 2 & ... \\
\hline\hline
Erasure & & & x & x & & ...\\
\hline
\end{tabular}
\label{table:erasure_ex}
\end{table}

\noindent{\bf Performance Metric.} The objective is to minimize the regret
\begin{equation}\label{eq:regret}
    R_T = \sum_{t=1}^{T} \max_{i\in [K]} \mu_i - \sum_{t=1}^{T} \mu_{\Tilde{a}_t}.
\end{equation}

\section{Motivation and Challenges}
\label{sec:prelim}
In this paper we consider a specific  agent operation, namely, the agent simply plays the most recently received action.
In this section, we argue that this is a reasonable choice,
by considering alternate 
simple strategies and explaining why they do not work well. We also discuss the technical challenges of our approach.  

\noindent{\bf Random Action.}
A very simple strategy could be, when an erasure occurs, for the agent to  uniformly at random select one of the $K$ actions to play. 
 The rewards seen by the learner will follow a new distribution with mean that is shifted from the pulled arm mean by a constant; in particular,\vspace{-5pt}
$$\vspace{-5pt}\mathbb{E}\left[r_t | a_t\right] = \left( 1 - \epsilon\right)\mu_{a_t} + \epsilon\mathbb{E}\left[r_{t}|\varepsilon_t\right] = \left( 1 - \epsilon\right)\mu_{a_t} + \frac{\epsilon}{K}\sum_{k=1}^K \mu_k,$$
where $\varepsilon_t$ is the event that erasure occurs at time $t$. Hence, the best arm does not change, the  gaps between arm means do not change, and the learner can identify the best arm and provide a good policy.
 However, as actions are erased in $\Omega(\epsilon T)$ iterations, the regret  becomes $\Omega(\epsilon T)$. 
 
\noindent{\bf Fixed Action.}
Very similar arguments hold if, when erasures occur, the agent always plays a fixed (predetermined) action $i$; unless $i$ happens to be the optimal action, the experienced regret will be linear.

\noindent{\bf Last Received Action.}
To see why the previous strategies fail, observe that  although the learner may have identified the best policy, this will not be consistently played. In this paper, we make the assumption that if an action is erased, the agent plays the last successfully received action. Thus if the optimal action is identified, it will be consistently played. 

We note that our selected agent strategy introduces memory and a challenge in the analysis since it creates a more complicated dependency between the action and erasures. For example, any of the previously sent actions by the learner has a non-zero probability of being played at the current iteration, where the probability changes with time and the sequence of previous actions. In the previous two strategies, the learner still observes a reward with fixed mean and can be treated as a valid MAB instance (although the optimal action might be different from the ground truth). The last received action strategy unfortunately changes the reward distribution and standard MAB analysis no longer holds.

\section{Repeat-the-Instruction Module}
\label{sec:alg1}
In this section we propose a scheme that works on top of any MAB algorithm to make it robust to erasures. Our scheme adds a form of repetition coding layer on top of the MAB algorithm operation. In particular, if the underlying MAB algorithm selects an action to play, our scheme sends the action chosen by the algorithm $\alpha$ times to the agent, and only associates the last received reward with the chosen action (hence the name Repeat-the-Instruction).  Since the agent plays the last received action, one successful reception within the $\alpha$ slots is sufficient to make the last reward sampled from the distribution of the chosen arm. Thus our algorithm, summarized in  Agorithm~\ref{alg1}, effectively partitions the $T$ rounds into $T/\alpha$ groups of rounds, where each group of $\alpha$ rounds is treated by the underlying bandit algorithm as one time slot.
The next theorem describes what is the resulting performance. 
 
\begin{algorithm}
\caption{\label{alg1} Repeat-the-Instruction Module}
\begin{algorithmic}[0] 
\State{\textbf{Input:} $\alpha, \text{ALG}$}
\For{$t \gets 1,..., T$}
    \State{\textbf{Learner:}}
    \State{$~~$1) $a_t = \text{ALG}(\frac{t-1}{\alpha}+1)$ \textbf{if} $(t\equiv 1\mod \alpha)$\\
    \qquad\qquad \textbf{else} $a_t = a_{t-1}$}
    \State{$~~$3) observe $r_t$ (corresponding to $\tilde{a}_t$) \textbf{if} $(t\equiv 0\mod \alpha)$}

    \State{\textbf{Agent:}}
    \State{$~~$2) play $\tilde{a}_{t-1}$ \textbf{if} \text{erasure} \textbf{else} play $a_{t}$}.\\ \qquad \qquad $\tilde{a}_0$ is initialized uniformly at random.
\EndFor
\end{algorithmic}
\end{algorithm}

\begin{theorem}\label{thm:main}
    Let {ALG be a MAB algorithm  with expected regret upper bounded by $R^{\text{ALG}}_T(\{\Delta_i\}_{i=1}^K)$ for any  instance with gaps $\{\Delta_i\}_{i=1}^K$}. For $\alpha = \lceil 2 \log T / \log( \frac{1}{\epsilon})\rceil$, using Repeat-the-Instruction  on top of ALG achieves an expected regret $\mathbb{E}[R_T]$   \begin{equation}
     \mathbb{E}[R_T] \leq 2\alpha R^{\text{ALG}}_{\lceil \frac{T}{\alpha}\rceil}(\{\Delta_i\}_{i=1}^K)+\alpha+1,
     \end{equation}
     where the expectation is over the randomness in the MAB instance, erasures, and algorithm.
\end{theorem}
\textit{Proof.}
We use ``run" to refer to the $\alpha$ rounds that repeat each action dictated by ALG.
    Let $G$ be the event that all  runs contain at least one round with no erasure. 
    {Conditioned on $G$, the maximum number of consecutive plays for an action due to erasures is $2\alpha - 1$ (across two runs, when the action transmitted at the first round of the first run is not erased, while the first $\alpha-1$ rounds of the next run are erased). Hence, we have that 
    \begin{align}\label{eq:bnd-agent-0}
        \mathbb{E}\left[R_T | G \right] &\leq 2\alpha \mathbb{E}[\sum_{i\equiv 1\text{ mod } \alpha,\ \alpha \leq t\leq T} \mu_{a^\star}-\mu_{a_t}| G ]+\alpha,
    \end{align}
    where $a^\star$ is the optimal arm, the second term bounds the regret for the first $\alpha$ iterations. Note that, conditioned on $G$, 
     if $t\equiv 0 \text{ mod }\alpha$ with  $t\geq \alpha$, we have that $\tilde{a}_{t}=a_{t-\alpha+1}$, that is,   
     the reward $r_t$  is sampled from arm $a_{t-\alpha+1}$. Indeed in Algorithm~\ref{alg1}  $r_t$ is the reward associated with arm $a_{t-\alpha+1}$ for $t\equiv 0 \text{ mod }\alpha, t\geq \alpha$. We observe that as erasures occur independently of the bandit environment and learners actions, we have that for $t\equiv 0 \text{ mod }\alpha, t\geq \alpha$, any set $A\subseteq \mathbb{R}$ and action $a_{t-\alpha+1}$, we have that $\mathbb{P}[r_t\in A|G,a_{t-\alpha+1}]=\mathbb{P}[r_t\in A|\tilde{a}_{t}=a_{t-\alpha+1}]$ (note that $G$ only depends on erasures). In particular, conditioned on $G$, the algorithm ALG receives rewards generated from a bandit instance with the same reward distributions as the original instance, hence, the same gaps, and time horizon $\lceil \frac{T}{\alpha} \rceil$. Hence, its regret is upper bounded by $R^{\text{ALG}}_{\lceil \frac{T}{\alpha}\rceil}(\{\Delta_i\}_{i=1}^K)$. Substituting in \eqref{eq:bnd-agent-0} we get that 
    \begin{equation}\label{eq:bnd-agent}
        \mathbb{E}\left[R_T | G \right] \leq 2\alpha R^{\text{ALG}}_{\lceil \frac{T}{\alpha}\rceil}(\{\Delta_i\}_{i=1}^K)+\alpha.
    \end{equation}}
    We finally upper bound the probability of the event $G$. Let $G_i$ denote the event that run $i$ contains at least one slot with no erasure. The probability of $G^C$ can be bounded as
    \begin{equation}
        \mathbb{P}[G^C] \stackrel{(1)}{\leq} \sum_{i=1}^{\lceil T/\alpha\rceil}\mathbb{P}[G_i^c] =\sum_{i=1}^{\lceil T/\alpha\rceil} \epsilon^{\alpha}\leq \sum_{i=1}^{\lceil T/\alpha \rceil} \frac{1}{T^2}\leq 1/T,
    \end{equation}
    where $(1)$ follows by the union bound. From \eqref{eq:bnd-agent}, we get that
    \begin{align*}
        \mathbb{E}\left[R_T\right]&\leq 2\alpha R^{\text{ALG}}_{\lceil \frac{T}{\alpha}\rceil}(\{\Delta_i\}_{i=1}^K)+\alpha+T \mathbb{P}[G^C]\nonumber \\
        &\leq 2\alpha R^{\text{ALG}}_{\lceil \frac{T}{\alpha}\rceil}(\{\Delta_i\}_{i=1}^K)+\alpha+1. \qquad \blacksquare
    \end{align*}
\begin{corollary}
    For $\alpha = \lceil 2\log T / \log( \frac{1}{\epsilon})\rceil$,  if $R_T^{\text{ALG}}$ is the worst-case expected regret of ALG, then  
    \begin{equation}
    \mathbb{E}[R_T]\leq R_T^{\text{ALG}}/\sqrt{1-\epsilon}.\end{equation}
    \end{corollary}
    This follows from Theorem~\ref{thm:main} by  observing that $\log(1/\epsilon)=\Omega(1-\epsilon)$, and that the worst-case regret for any algorithm is $\Omega(\sqrt{KT})$.
    The next corollary  follows by substituting the regret bound of the UCB algorithm \cite{Auer2002, Lai1987AdaptiveTA} in Theorem~\ref{thm:main}.
\begin{corollary}
    For $\alpha = \lceil 2\log T / \log( \frac{1}{\epsilon})\rceil$, the proposed algorithm with the UCB algorithm \cite{Auer2002, Lai1987AdaptiveTA} achieves a gap-dependent regret bounded by
    $\mathbb{E}[R_T]\leq c \alpha \sum_{i:\Delta_i> 0}\frac{\log T}{\Delta_i},$
    and a worst-case regret bounded by
     $\mathbb{E}[R_T]\leq c \sqrt{T K\log{T}/(1-\epsilon)}$,
    where $c$ is a universal constant.
\end{corollary}
\noindent{\bf Observation.} Note that our proposed module achieves the optimal dependency of the regret on $T$ and $K$. However, we show next that the multiplicative dependency on $\epsilon$ is suboptimal by providing an algorithm with a regret bound that has additive dependency on $\epsilon$. This effect is small for small values of $\epsilon$ but becomes significant for $\epsilon$ approaching $1$.}

\section{The Lingering {SAE (L-SAE)} Algorithm}\label{sec:alg2}
The intuition behind the repeat-the-instruction algorithm is that we do not frequently switch between arms - and thus playing the last successfully received action often coincides with playing the desired action. In this section, we propose ``Lingering {Successive Arm Elimination" (L-SAE)}, an algorithm that by design does not frequently switch arms, and evaluate its performance; in the next section, we prove a lower bound establishing {L-SAE} is order optimal.

{L-SAE} builds on the Successive Arm Elimination (SAE) algorithm \cite{auer2010ucb}. SAE works in batches of exponentially growing length, where at the end of each batch the arms that appear to be suboptimal are eliminated. In batch $i$, each of the surviving arms is pulled $4^i$ times. We add two modifications to SAE to make it robust to erasures. First, we do not use the frequent arm switches  of the first batches, when $4^i$ is small. {Instead, in the first batch, the algorithm pulls each arm $4\alpha$ times. Then, the number of pulls for surviving arms in a batch is $4$ times that of the previous batch.}
Second, we ignore the first half of the samples for each arm. This ensures that all the chosen rewards are picked from the desired arm with high probability, as the probability of half the samples being erased is small. Given these modifications, we also update the arm elimination criterion  to account for the higher variance in the mean estimates. In particular, the algorithm starts with $A=[K]$ as the good arms set and at the end of batch $i$, the algorithm eliminates arms with empirical mean that is away by more than $\sqrt{\log(KT)/(\alpha 4^{i-2})}$ from the empirical mean of the arm that appears to be best. The pseudo-code of the algorithm is provided in Algorithm~\ref{alg:SAE}.
\begin{algorithm}
  \caption{Lingering {SAE} Algorithm }
\label{alg:SAE}  
\begin{itemize}
    \item Initialize: set of good arms $A=[K]$, batch index $i=1$. 
    \item For batch $i$:
    \begin{itemize}
        \item Pull each arm in $A$, $M_i=\alpha 4^i$ times to receive rewards $r^a_1,...,r^a_{M_i}$ for arm $a\in A$.
        \item Update means: $\mu_a^{(i)}={\sum_{j=M_i/2+1}^{M_i}r^a_j}/({M_i/2})\ \forall a\in A$.
        \item $A\gets \{a\in A| \max_{\tilde{a}\in A}\mu_{\tilde{a}}^{(i)}-\mu_a^{(i)} \leq 4\sqrt{{\log (KT)}/{M_i}}\}$.
        \item $i\gets i+1$.
    \end{itemize}
\end{itemize}
\end{algorithm}
\begin{theorem}\label{thm:optimal}
    For $\alpha = \lceil 2 \log T / \log( \frac{1}{\epsilon})\rceil$, Algorithm~\ref{alg:SAE} achieves a regret that is bounded by
    $$R_T\leq c\left(\frac{K\log T}{1-\epsilon}+\sum_{i:\Delta_i>0}\frac{\log T}{\Delta_i}\right)$$ with probability at least $1-1/T$, and for a constant $c$.
\end{theorem}
\noindent{\textit{Proof.}
Let $G$ be the good event that the second half of all arm pulls in all batches are from the correct (desired) arm. Note that $G$ does not occur when there is a batch $i$ and an arm $j$ such that all half of arm $j$ pulls in batch $i$ coincide with an erasure. As since in any batch each arm is pulled at least $2\alpha$ times we have that\vspace{-5pt}
\begin{align}
    \mathbb{P}[G] = 1-\mathbb{P}[G^C]\geq 1-K\log (T) \epsilon^\alpha\geq 1-0.25/T.
\end{align}
{We notice that erasures are independent of the bandit environment, and learner actions; and $G$ only depends on erasures. Hence, conditioned on $G$ and the picked arm $a$, the second half of the rewards $\{r_j^a\}_{j=M_i/2+1}^{M_i}$ are picked from arm $a$ and they follow the original reward distribution of arm $a$. Hence, as the rewards are only supported on $[0,1]$, we have that conditioned on $G,a$, the reward $r_j^a, j>M_i/2$ is $1/4$-subGaussian with mean $\mu_a$.} By concentration of sub-Gaussian random variables, conditioned on $G$, we also have that the following event
$$G'=\{|\mu_a^{(i)}-\mu_a|\leq 2\sqrt{{\log (KT)}/{M_i}}\forall a\in A_i\forall i\in [\log T]\},$$
where $M_i$ is the number of pulls for surviving arms in batch $i$, occurs with probability at least $1-0.25/T$. Hence, we have that\vspace{-5pt}
$$\mathbb{P}[G\cap G']\geq (1-0.25/T)^2\geq 1-1/T.$$
In the remaining part of the proof we condition on the event $G \cap G'$. By the elimination criterion in Algorithm~\ref{alg:SAE}, and assuming $G \cap G'$, the best arm will not be eliminated. This is because the elimination criterion will not hold for the best arm as\vspace{-5pt}
\begin{align}
    \mu_a^{(i)}-\mu_{a^\star}^{(i)}&\leq \mu_a-\mu_{a^\star}+4\sqrt{\log(KT)/M_i}\nonumber \\
    &\leq 4\sqrt{\log(KT)/M_i}\forall a \forall i.
\end{align}
Now consider an arm with gap $\Delta_a>0$ and let $i$ be the smallest integer for which $4\sqrt{\log(KT)/M_i}< \frac{\Delta_a}{2}$. Then, we have that 
\begin{align}
    \mu_{a^\star}^{(i)}-\mu_a^{(i)}&\geq \mu_{a^\star}-\mu_a-4\sqrt{\log(KT)/M_i}>\Delta_a - \frac{\Delta_a}{2}\nonumber \\
    &>4\sqrt{\log(KT)/M_i}.
\end{align}
Hence, arm $a$ will be eliminated before the start of batch $i+1$. We also notice that from $4\sqrt{\log(KT)/M_i}< \frac{\Delta_a}{2}$, the value of $i$ can be bounded as
\begin{equation}
    i\leq \max\{1, \log_4(\frac{65\log(KT)}{\alpha \Delta_i^2})\},
\end{equation}

By the exponential increase in the number of pulls of each arm, we get that until eliminated, arm $a$ will be pulled by the learner at most $$T_a(T)\leq \sum_{j=1}^{i} 4^j\alpha\leq 4^{i+1}\leq c(\alpha+\frac{\log(KT)}{\Delta_a^2}),$$
for some absolute constant $c>0$. We also notice that on event $G$, the agent will pull arm $i$ at most $4T_i(T)$ times. This results in a regret that is at most $c (\alpha+\log(KT)/\Delta_a)$. Summing the regret over all arms, we get that conditioned on $G\cap G'$ we have that
$$R_T\leq c\left(\frac{K\log T}{\log(1/\epsilon)}+\sum_{a:\Delta_a>0}\frac{\log(KT)}{\Delta_a}\right).$$
The proof is concluded by noticing that $\log(1/\epsilon)=O(1-\epsilon)$.}
$\blacksquare$\\
The previous theorem directly implies the following worst-case regret bound
 $R_T\leq c\left(\frac{K\log T}{1-\epsilon}+\sqrt{KT}\right).$

{\section{Lower Bound}\label{sec:lower-bnd}
We here prove a lower bound that matches the upper bound in Theorem~\ref{thm:optimal} up to $\log$ factors. 
A lower bound of $\Omega(\sum_{i:\Delta_i>0}\frac{1}{\Delta_i})$ on the gap-dependent regret is already provided in \cite{lai1985asymptotically} and a lower bound on the worst-case regret of $\Omega(\sqrt{KT})$ is provided in \cite{auer1995gambling}. Thus it suffices to prove a lower bound of $\Omega(K/(1-\epsilon))$ which we provide next
 in Theorem~\ref{th3}.
\begin{theorem} \label{th3}
    Let $T\geq \frac{K}{4\log(1/\epsilon)}$, and $\epsilon\geq 1/2$. Assuming the agent operation in Section~\ref{sec:model}, for any policy $\pi$, there exists a $K$-armed bandit instance $\nu$ such that
    \begin{equation}
        \mathbb{E}[R_T(\pi, \nu)]\geq c\frac{K}{1-\epsilon},
    \end{equation}
    where $\mathbb{E}[R_T(\pi, \nu)]$ is the expected regret of the policy $\pi$ over the instance $\nu$ and $c$ is a universal constant.
\end{theorem}
\noindent\textit{Proof.}
We consider $K$ bandit instances, each with $K$ arms, where in instance $\nu_i$   the means of the reward distributions are
\begin{equation}
    \mu_j^{(i)} = \mathbf{1}\{j=i\}, j=1\ldots K
\end{equation}
with $\mathbf{1}$ the indicator function. We assume a noiseless setting where in instance $\nu_i$ picking arm $j$ results in reward $\mu^{(i)}_j$ almost surely. We consider two events:\\
$E_i$ indicates that the first $1/\log(1/\epsilon)$ pulls of arm $i$ are erased;\\ 
$E'_i=\{\tilde{a}_0\neq i\}$  indicates that the agent in the first round, if the first transmitted arm is erased, does not select to pull arm $i$ (recall that if an erasure occurs at the first round the agent randomly selects an action according to some distribution). 
{We note that $E_i, E'_i$ are independent events since erasures are independent of the agent operation.} 

We will show that there is no policy $\pi$ that can make $\mathbb{E}[R_T(\pi, \nu_i)|E_i\cap E'_i]$ to be small for all $i$. We first note that there are at least\footnote{We assume for simplicity that $K$ is divisible by $4$. The proof easily generalizes by taking the floor in divisions, which only affects the constants.}  $K/2$ arms with $\mathbb{P}[E'_i]\geq 1/2$. Pick a set $I\subseteq [K]$ with $|I|=K/2$, and $\mathbb{P}[E'_i]\geq 1/2$ $\forall i\in I$.
Now define the event $\mathcal{P}_i$:\\ 
    $\mathcal{P}_i$:  arm  $i$ is picked no more than $\frac{1}{\log(1/\epsilon)}$  times by the
    learner in the first  $\frac{K}{4\log(1/\epsilon)}$  rounds.\\
{We next consider the minimum worst case probability of  $\mathcal{P}_i$ conditioned on $E_i\cap E'_i$ under the distribution induced by instance $\nu_i$; namely, 
\begin{equation}\label{eq:lb-min-max}
    \min_{\pi}\max_{i\in I}\mathbb{P}_{\nu_i}[\mathcal{P}_i|E_i\cap E'_i].
\end{equation}
Let the set $A$ denote the indices of arms for which $\mathcal{P}_i$ holds, i.e., $A=\{i\in I|\mathcal{P}_i\}$, and let $B=I/A$. We have that $A,B$ are disjoint and their union is the set $I$. Moreover, the quantity in \eqref{eq:lb-min-max} can be rewritten as
\begin{equation}
    \min_{\pi}\max_{i\in I}\mathbb{P}_{\nu_i}[i\in A|E_i\cap E'_i].
\end{equation}
We make two observations: (i) after $\frac{K}{4\log(1/\epsilon)}$ rounds, there exist at least $\frac{|I|}{2}=\frac{K}{4}$ arms in $I$ for which $\mathcal{P}_i$ holds (hence, $|A|\geq |I|/2$ and $|B|\leq |I|/2$); and (ii)
if for instance $\nu_i$, conditioned on $E_i\cap E'_i$,  arm $i$ is picked less than $1/\log(1/\epsilon)$ times by the learner, then all the rewards received by the learner will be zeros, thus no information will be provided to the policy during the first  $\frac{K}{4\log(1/\epsilon)}$ rounds.\\
From observation (ii) and the fact that the result of whether $i\in A$ or not, does not change after the first reward feedback that is $1$, it follows that the optimal value for \eqref{eq:lb-min-max} does not change if the learner does not observe rewards. 
Hence, the learner can proceed assuming that all previous rewards are zeros. As a result, the learner can decide on the arms to pull in the first $K/(4\log(1/\epsilon))$ slots ahead of the time (i.e., at $t=1$; since no feedback is required); equivalently, the learner can divide the set $I$ into two sets $A$ and $B$ (possibly in a random way) ahead of the time with $|B|\leq |I|/2$ (from observation $(i)$). As the learner decides on $A$ ahead of the time, the probability of $i\in A$ does not depend on the instance, reward outcomes and $\tilde{a}_0$, hence, we can simply denote $\mathbb{P}_{\nu_i}[i\in A|E_i\cap E'_i]$ as $\mathbb{P}[i\in A]$. This shows that, the problem of minimizing $\max_{i\in I}\mathbb{P}_{\nu_i}[\mathcal{P}_i|E_i\cap E'_i]$ is equivalent to partitioning the set $I$ into two sets $A$ and $B$ (with $A$ containing the indices where $\mathcal{P}_i$ holds), $|A|\leq |I|/2$ and with the goal of minimizing $\max_{i\in I}\mathbb{P}[i\in A]$.}
It is easy to see that the minimum value for $\max_{i\in I}\mathbb{P}[i\in A]$, and similarly, $\max_{i\in I}\mathbb{P}_{\nu_i}[\mathcal{P}_i|E_i\cap E'_i]$, is at least $1/2$ as $\sum_{i=1}^{|I|}\mathbb{P}[i\notin A]\leq |I|/2$.

{Hence, for any policy $\pi$, there is an instance $\nu_i, i\in I$ such that conditioned on $E_i\cap E'_i$, arm $i$ is picked no more than $1/\log(1/\epsilon)$ times by the learner in the first $K/2\log(1/\epsilon)$ rounds with probability at least $1/2$.
But for instance $\nu_i$, whenever arm $i$ is not pulled we incur a regret of value $1$, we get that there is $i\in I$ such that $\mathbb{P}\left[R_T(\pi,\nu_i)\geq \frac{K}{4\log(1/\epsilon)}|E_i\cap E'_i\right]\geq 1/2$.} Then, we have that $\mathbb{E}[R_T(\pi,\nu_i)|E_i\cap E'_i]\geq c\frac{K}{\log(1/\epsilon)}$. By non-negativity of regret, we have that
\begin{align}
    \mathbb{E}[R_T(\pi,\nu_i)]&\geq c\frac{K}{\log(1/\epsilon)}\mathbb{P}[E_i\cap E'_i]\nonumber \\
    &\stackrel{(i)}{=} c\frac{K}{\log(1/\epsilon)}\mathbb{P}[E_i]\mathbb{P}[E'_i]\nonumber \\
    &\stackrel{(ii)}{\geq} c/2 \frac{K}{\log(1/\epsilon)}\mathbb{P}[E_i],
\end{align}
where $(i)$ follows from the fact that $E_i,E'_i$ are independent, and $(ii)$ follows since $i\in I$ and thus $P(E'_i)\geq \frac{1}{2}$. Moreover, $\mathbb{P}[E_i]=\epsilon^{1/\log(1/\epsilon)}=e^{-1}$ and thus $\mathbb{E}[R_T(\pi,\nu_i)]\geq c\frac{K}{\log(1/\epsilon)}$. The proof is concluded by observing that $\log(1/\epsilon)=O(1-\epsilon)$ for $\epsilon\geq 1/2$.
$\hfill{\blacksquare}$
}